\documentclass{article}

\usepackage[final]{neurips_2024}

\usepackage[utf8]{inputenc} 
\usepackage[T1]{fontenc}    
\usepackage{hyperref}       
\usepackage{url}            
\usepackage{booktabs}       
\usepackage{amsfonts}       
\usepackage{nicefrac}       
\usepackage{microtype}      
\usepackage{xcolor}         
\usepackage{listings}
\usepackage{pgfplots}
\usepackage{subcaption}
\usepackage{float}
\usepackage{geometry}
\pgfplotsset{compat=1.18}
\usepackage{amsmath}

\title{Finetune-RAG: Fine-Tuning Language Models to Resist Hallucination in Retrieval-Augmented Generation}

\author{%
  Zhan Peng Lee \\
  Pints AI Labs\\
  \texttt{zhanpeng.lee@pints.co} \\
  \And
  Andre Lin\thanks{Work was done during an internship at Pints AI} \\
  Pints AI Labs \\
  \texttt{andre\_lin@u.nus.edu} \\
  \texttt{andrelim444@gmail.com} \\
  \And
  Calvin Tan \\
  Pints AI Labs\\
  \texttt{calvin@pints.co} \\
}

\begin{document}

\maketitle

\begin{abstract}
Retrieval-Augmented Generation (RAG) has emerged as a powerful framework to improve factuality in large language models (LLMs) by grounding their outputs in retrieved documents. However, ensuring perfect retrieval of relevant information remains challenging, and when irrelevant content is passed downstream to an LLM, it can lead to hallucinations. In this work, we propose Finetune-RAG, a simple and effective fine-tuning approach that features the first-of-its-kind RAG training dataset constructed to mimic real-world imperfections. Experimental results show that Finetune-RAG improves factual accuracy by 21.2\% over the base model. We also propose Bench-RAG, an LLM-as-a-judge evaluation pipeline that stress tests models under realistic imperfect retrieval scenarios. Our codebase\footnote{\url{https://github.com/Pints-AI/Finetune-Bench-RAG}} and dataset\footnote{\url{https://huggingface.co/datasets/pints-ai/Finetune-RAG}} are fully open sourced for community use.
\end{abstract}

\section{Introduction}

Large Language Models (LLMs) have demonstrated remarkable capabilities across a wide range of natural language processing tasks \citep{wang2023codet5opencodelarge, rozière2024codellamaopenfoundation, cui2025multilingualmachinetranslationopen, yasunaga2022qagnnreasoninglanguagemodels, liu-etal-2024-learning}. However, their tendency to "hallucinate", that is, to produce fluent but factually incorrect information, remains a persistent challenge \citep{li2024dawndarkempiricalstudy, duan2024llmsknowhallucinationempirical, zhang2023sirenssongaiocean}, particularly in high-stakes domains such as healthcare, law, and finance \citep{agarwal2024medhaluhallucinationsresponseshealthcare, Dahl_2024, kang2023deficiencylargelanguagemodels}. To address this, {\bf Retrieval-Augmented Generation (RAG)} has become a popular solution. Instead of relying solely on parametric memory, RAG systems retrieve external documents and condition the model’s response on this evidence.

In practice, retrieval accuracy in RAG is far from flawless. Retrieved documents may be outdated, misleading, or topically adjacent but factually incorrect. These errors can propagate downstream, leading models to blend inaccurate context into fluent but false answers. This is especially concerning in domains such as law, compliance, financial reporting, or medicine, where mistakes can have wide-ranging repercussions.

Most prior work has addressed this issue from the retrieval perspective, focusing on improving retrievers, reranking mechanisms, or applying filtering heuristics \citep{Sawarkar_2024, dong2024dontforgetconnectimproving, zhou2025openragoptimizingragendtoend}. In contrast, relatively little attention has been given to improving the {\bf model's ability to resist using the incorrect information}.

In this paper, we introduce {\bf Finetune-RAG}, a method that directly targets hallucination by fine-tuning the model with imperfect RAG samples that mimic real-world retrieval scenarios. We constructed a diverse dataset covering legal documents, scientific literature, books, and web data, each paired with a plausible but fictitious counterpart. We then fine-tune instruction-tuned LLMs, specifically Meta's Llama 3.1-8B-Instruct \citep{grattafiori2024llama3herdmodels}, on this dataset using two prompt variants: a {\bf Baseline format} and a {\bf Structured XML} variant. This setup allows us to assess generalization and prompt sensitivity. To our knowledge, Finetune-RAG provides the first RAG dataset of its kind, as existing RAG finetuning datasets implicitly assume perfect information retrieval, and mostly focus only the LLM's ability to extract coherent answers from relevant chunks.

Our key insight is that LLMs struggle to identify contextual clues that are obvious to the human eye, such as financial reports from a similarly named company or outdated information based on dates indicated by document metadata. Through fine-tuning models with a controlled mixture of true and false context placed alongside, we teach them to ground their answers exclusively in the reliable information provided.

We evaluated the effectiveness of Finetune-RAG using {\bf Bench-RAG}, a custom benchmarking suite we have created  that leverages {\bf GPT-4o} \citep{openai2024gpt4o} as an automated judge to assess the accuracy, relevance, helpfulness and depth of the LLM response. Our results show that Finetune-RAG substantially improves factual correctness while maintaining output quality across other dimensions, demonstrating that generation-time defenses are a viable complement to improved retrieval.

Our contributions are as follows:

\begin{itemize}
    \item \textbf{Fine-tuning Approach.} We propose a novel fine-tuning strategy for RAG systems that teaches models to ignore misleading context and generate answers based solely on factual input.
    
    \item \textbf{Training Dataset.} We release a curated, multi-domain dataset designed for hallucination resistance training, with both factual and fictitious content.

    \item \textbf{Evaluation Setup.} We benchmark the effectiveness of our approach using GPT-4o-based evaluations and show significant gains in factual accuracy without compromising helpfulness or relevance.

    \item \textbf{Open-source release.} We make our code, models, dataset, and evaluation framework publicly available to facilitate further research. They can be accessed in our open-source repository\footnote{\url{https://github.com/Pints-AI/Finetune-Bench-RAG}} and dataset\footnote{\url{https://huggingface.co/datasets/pints-ai/Finetune-RAG}}.
\end{itemize}

By fine-tuning LLMs on RAG examples containing both factual and fictitious documents, we show that it is possible to build models that can reliably choose truth over noise. Our dataset reflects noisy, domain-diverse retrieval as encountered in practice, making it a strong foundation for stress-testing hallucination resistance in future RAG systems.

\section{Background}
\label{background}

\subsection{Retrieval-Augmented Generation}
Retrieval-Augmented Generation (RAG) augments large language models by incorporating external documents into the generation process. Rather than relying solely on the model's internal parameters, RAG retrieves relevant passages from a knowledge base and feeds them, along with the user query, into the model to guide its response \citep{zhou2024trustworthinessretrievalaugmentedgenerationsystems}.

A standard RAG system operates in two phases:
\begin{itemize}
    \item \textbf{Retrieval.} A retriever model selects the top-$k$ most relevant documents for a given query.

    \item \textbf{Generation.} A language model generates a response conditioned on both the query and the retrieved documents.
\end{itemize}
The appeal of RAG lies in its ability to dynamically access up-to-date or domain-specific information, which is especially useful in fast-changing or specialized fields. However, it also introduces new failure modes, particularly when the retrieval quality is imperfect \citep{barnett2024sevenfailurepointsengineering}.

\subsection{Hallucination in Language Models}

Hallucination refers to the phenomenon where language models produce outputs that are factually incorrect or unsupported by the input, resulting in unfaithful outputs \citep{rawte2023troublingemergencehallucinationlarge}. In RAG systems, hallucination can be especially problematic when the model is presented with a mixture of relevant and irrelevant (or even misleading) context. Even with carefully worded prompts, models can inadvertently "trust" incorrect sources and generate plausible but wrong answers \citep{yoran2024makingretrievalaugmentedlanguagemodels}.

Despite the presence of external context, most current models lack mechanisms to actively filter or ignore misleading information once it is included in the prompt \citep{shi2023largelanguagemodelseasily}. Finetune-RAG specifically targets this weakness by training models to develop this filtering capability.

\section{Related Works}
\label{related-works}

\subsection{Mitigating Hallucination with Synthetic Prompt Tuning}

SYNTRA \citep{jones2023teachinglanguagemodelshallucinate} reduces hallucinations in large language models by modifying the model's instructions rather than adjusting its internal weights. SYNTRA does this by attaching a small, trainable embedding vector to the system message, which acts as an additional instruction prefix. This vector is optimized using a synthetic task where hallucinations are easy to measure. For example, the model is prompted to return names starting with a specific letter from a visible list, and any incorrect or invented names are counted as hallucinations. By learning to avoid such mistakes in a controlled setting, the model can generalize to reduce hallucinations in downstream tasks. However, because SYNTRA focuses on modifying prompts and not the model's internal reasoning, it does not enable the model to distinguish between factual and misleading content, failing to address real-world RAG scenarios \citep{barnett2024sevenfailurepointsengineering}\citep{shi2023largelanguagemodelseasily}.

\subsection{Refusal-Aware Fine-Tuning}

\citet{zhang2024rtuninginstructinglargelanguage} propose a fine-tuning method, R-Tuning, that teaches language models to express uncertainty and decline to answer when a question falls outside their pre-trained knowledge. This is achieved by identifying questions the model answers incorrectly during training and appending an uncertainty statement such as “I am unsure” to those responses. The result is a model that behaves more conservatively and with improved confidence calibration. However, R-Tuning is designed for closed-book settings, where the model relies only on its internal knowledge without a RAG system.

\subsection{Constrained Reasoning with Decompose-and-Query (D\&Q)}

\citet{cao2023stepclosercomprehensiveanswers} propose the Decompose-and-Query (D\&Q) framework, which extends retrieval-augmented generation (RAG) by teaching language models to break down complex queries, retrieve relevant information using external tools, and generate answers based on a structured knowledge source. In particular, D\&Q introduces a curated question–answer (QA) base, which is a collection of verified QA pairs that the model consults during reasoning. This setup helps reduce hallucinations by constraining the model to reliable content and allowing it to backtrack when inconsistencies are detected.

However, the effectiveness of D\&Q depends strongly on the quality and coverage of its QA base. In practical RAG applications, where retrieved content can be noisy, ambiguous, or incomplete \citep{shi2023largelanguagemodelseasily}, relying on a fixed and curated source may become a limitation. Since the framework lacks mechanisms to dynamically assess the reliability of new information, it remains susceptible to hallucinations caused by misleading or inaccurate context.

\section{Methodology}
\label{methodology}

We introduce {\bf Finetune-RAG}, a fine-tuning method designed to train large language models (LLMs) to distinguish between correct and fictitious context within a Retrieval-Augmented Generation (RAG) setup. Unlike prior work that attempts to improve factuality by enhancing the retrieval phase, Finetune-RAG focuses on improving the model’s generation behavior when faced with imperfect or misleading inputs. Our core idea is to fine-tune the model using examples where both correct and incorrect information are explicitly presented to model, allowing it to learn the ability to sift out the correct information to use for its response.

\subsection{Problem Setup}
\label{problem-setup}

In a typical RAG system, the model is given a user query 
$q$ and a set of retrieved documents $\{d\textsubscript{1}, d\textsubscript{2}, ..., d\textsubscript{k}\}$ \citep{zhou2024trustworthinessretrievalaugmentedgenerationsystems}. When any of the documents is irrelevant or misleading, the model may generate incorrect responses \citep{yoran2024makingretrievalaugmentedlanguagemodels}.

In Finetune-RAG, we simulate this scenario during training by constructing prompts that include:

\begin{itemize}
    \item One correct (factual) document chunk $d\textsubscript{correct}$
    
    \item One fictitious (misleading) document chunk $d\textsubscript{fictitious}$

    \item A corresponding question $q$

    \item A reference answer $a$, written using only $d\textsubscript{correct}$ as the reference
\end{itemize}

The model is then trained using supervised fine-tuning to produce the answer $a$ despite having access to both $d\textsubscript{correct}$ and $d\textsubscript{fictitious}$ in the input.

In Bayesian modeling, we can think of the task as a conditional generation problem where the goal is to maximize the probability of generating a truthful answer $a$ given a question $q$ and a mixed set of contexts (some correct $d_{\text{correct}}$, some fictitious $d_{\text{fictitious}}$).

We aim to model:

\begin{equation}
P(a \mid q, d_{\text{correct}}, d_{\text{fictitious}})
\end{equation}

However, this is the observed conditional probability, and what we want the model to learn is to ignore $d_{\text{fictitious}}$ and generate the answer as if conditioned only on $d_{\text{correct}}$. So our training objective is to align to the following idealized posterior:

\begin{equation}
P^*(a \mid q, d_{\text{correct}}, d_{\text{fictitious}}) \rightarrow P(a \mid q, d_{\text{correct}})
\end{equation}

In other words, even though the model receives both correct and fictitious information, it must assign zero (or negligible) attention/mass to $d_{\text{fictitious}}$ during decoding.

\subsection{Prompt Construction}

Each training example in Finetune-RAG is processed to include a \textbf{system message} and a \textbf{user message}, following the standard instruction-tuning format \citep{ouyang2022traininglanguagemodelsfollow} used in chat-style language models. The system message defines the behavior of the assistant, while the user message provides the question along with correct and fictitious information.

\subsubsection{System Message}

The system message is consistent in all training examples. It instructs the assistant to rely solely on the provided context and discourages the use of prior knowledge or hallucination:

\begin{verbatim}
"Some information is retrieved from the database as provided based on the
user’s question. The assistant is to answer the question to the best of
his/her ability, using only the information provided. The assistant must
not add his/her own knowledge."
\end{verbatim}

\subsubsection{User Message}
\label{user-message}

To help the model distinguish between factual and fictitious context more effectively, we explore the use of XML-like \citep{bray1998xml} structured input. We hypothesize that introducing a consistent and explicit hierarchy, where document chunks are clearly labeled and separated, can make it easier for the model to parse and evaluate different sources of information. This is especially important in RAG settings, where hallucinations often result from the model blending or misattributing content across documents. Our approach aligns with findings from recent work such as StructRAG \citep{li2024structragboostingknowledgeintensive} and SRAG \citep{lin2025sragstructuredretrievalaugmentedgeneration}, which demonstrates that task-specific structured representations such as tables or graphs can significantly improve the performance of LLMs on knowledge-intensive reasoning tasks. Our use of XML aims to impose syntactic clarity and boundary enforcement at the input level.

To test this, we compare two user message formats: an unstructured {\bf Baseline Format} and a structured {\bf XML Format}. Both present a question along with two document chunks, one factual and one fictitious, but differ in how the information is presented. Refer to Section \ref{prompt-structure-ablation} for the exact prompt structure.

\section{Experimental Setup}
\label{experimental-setup}

\subsection{Model}

We fine-tuned Meta's Llama 3.1–8B-Instruct \citep{grattafiori2024llama3herdmodels}, an instruct-tuned model that supports chat-style interaction and long context windows. We adapt the system and user message formatting based on the chosen prompt structure described in Section \ref{user-message}.

\subsection{Dataset and Preprocessing}

Our dataset contains a total of 1,653 examples from diverse domains, such as legal documents, scientific papers, news articles, and technical reports. For the complete structure of each example in the dataset, refer to Annex \ref{appendix:dataset-example-format}.

Each example is formatted in both the baseline and XML structures. The dataset is then partitioned into training (80\%), validation (10\%), and test (10\%) sets.

\subsection{Hyperparameters}

We selected hyperparameter values that balance model performance with computational efficiency. Refer to Table \ref{table:fine-tuning-hyperparameters} for the complete set of hyperparameters used.

\begin{table}[ht]
    \caption{Fine-tuning hyperparameters used on Llama 3.1-8B-Instruct}
    \label{table:fine-tuning-hyperparameters}
    \centering
    \begin{tabular}{ll}
        \toprule
        \textbf{Parameter} & \textbf{Value} \\
        \midrule
        Steps           & 20            \\
        Batch size      & 64           \\
        Learning rate   & 2e-5         \\
        Warmup ratio    & 0.1          \\
        LR Scheduler    & Cosine decay \\
        Optimizer       & AdamW        \\
        $\beta_1$       & 0.9          \\ 
        $\beta_2$       & 0.95         \\ 
        Weight decay    & 0.1          \\
        Mixed precision & BF16         \\
        \bottomrule
    \end{tabular}
\end{table}

\subsection{Checkpoints and Reproducibility}
\label{checkpoints}

We have released the model checkpoints fine-tuned with both Baseline\footnote{
\begin{minipage}[t]{\dimexpr\linewidth-3em}
\sloppy
\url{https://huggingface.co/pints-ai/Llama-3.1-8B-Instruct-RAG_Baseline_tuned-1} \\
\url{https://huggingface.co/pints-ai/Llama-3.1-8B-Instruct-RAG_Baseline_tuned-2}
\end{minipage}
} and XML\footnote{
\begin{minipage}[t]{\dimexpr\linewidth-3em}
\sloppy
\url{https://huggingface.co/pints-ai/Llama-3.1-8B-Instruct-RAG_XML_tuned-1} \\
\url{https://huggingface.co/pints-ai/Llama-3.1-8B-Instruct-RAG_XML_tuned-2}
\end{minipage}
} formats on HuggingFace. Each prompt structure has two repositories, and each repository contains five checkpoints, totaling 10 checkpoints each.

\section{Evaluation}
\label{evaluation}

We evaluate Finetune-RAG's ability to generate factually accurate answers when presented with both correct and fictitious context. Our evaluation framework focuses on measuring whether the model is able to \textit{selectively use only the correct information}, and we assess output quality across four key dimensions.

\subsection{Bench-RAG}
\label{bench-rag-setup}

We adopt a custom benchmarking pipeline, namely {\bf Bench-RAG}, using {\bf GPT-4o} model \citep{openai2024gpt4o} in a LLM-as-a-judge inspired by prior work\citep{zheng2023judgingllmasajudgemtbenchchatbot, gu2025surveyllmasajudge, li2025generationjudgmentopportunitieschallenges}. Using structured prompts to elicit consistent evaluations for each model output, we measure:

\begin{itemize}
    \item \textbf{Accuracy}: A binary metric indicating whether the generated answer is factually correct and based solely on the correct chunk. (True/False)
    \item \textbf{Helpfulness}: A score from 1 to 10 assessing how useful the answer is in addressing the user's question.
    \item \textbf{Relevance}: A score from 1 to 10 measuring how relevant the content is to the query.
    \item \textbf{Depth}: A score from 1 to 10 reflecting the level of detail or insight present in the answer.
\end{itemize}

Each generated output is rated using a structured prompt format, which requests scores across these categories and a brief justification. Refer to Appendix \ref{appendix:bench-rag-prompt-structure} for the full structure. This methodology draws from recent research demonstrating that LLMs can align closely with human preferences when prompted properly, achieving high inter-rater agreement, i.e. multiple evaluators provide consistent ratings for the same outputs \citep{gu2025surveyllmasajudge, li2025generationjudgmentopportunitieschallenges}.

\subsection{Checkpoints Evaluated}

For each prompt structure, we evaluate all 10 model checkpoints saved during training (see Section \ref{checkpoints}). These checkpoints represent the model's learning trajectory over the course of a single fine-tuning epoch. At each checkpoint, we generate answers to the test dataset questions using both the correct context $d\textsubscript{correct}$ and the fictitious context $d\textsubscript{fictitious}$. The generated answers are then submitted to the evaluator for scoring.
Refer to Appendix \ref{appendix:evaluation-system-message} and \ref{appendix:evaluation-user-message} for the structure of the prompt used for evaluation.

\subsection{Results}

We report quantitative results from our fine-tuning experiments scored across 4 dimensions: {\bf factual accuracy, helpfulness, relevance, and depth}. Evaluation was performed using GPT-4o \citep{openai2024gpt4o} as an LLM judge, as described in Section~\ref{bench-rag-setup}. We then aggregate the scores of each sequence in the test dataset to derive the final evaluation result for each checkpoint:

\begin{equation}
\bar{Accuracy} = \left( \frac{1}{n_{test}} \sum_{i=1}^{n_{test}} 1[Accuracy_i = True] \right) \times 100\%
\end{equation}
\begin{equation}
\bar{Helpfulness} = \frac{1}{n\textsubscript{test}} \sum_{i=1}^{n\textsubscript{test}} Helpfulness_i
\end{equation}
\begin{equation}
\bar{Relevance} = \frac{1}{n\textsubscript{test}} \sum_{i=1}^{n\textsubscript{test}} Relevance_i
\end{equation}
\begin{equation}
\bar{Depth} = \frac{1}{n\textsubscript{test}} \sum_{i=1}^{n\textsubscript{test}} Depth_i
\end{equation}

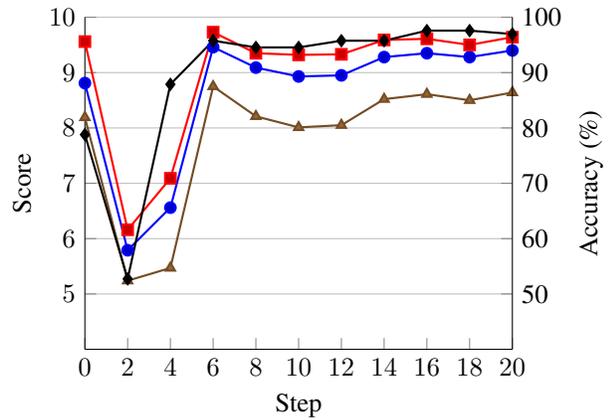
\begin{figure}[H]
    \caption{Evaluation results across training steps (Baseline format). Accuracy is plotted on the right y-axis, and other metrics use the left y-axis.}
    \label{figure:baseline-results}
    \centering
    \begin{minipage}{0.45\textwidth}
        \centering
        \begin{tabular}{ccccc}
            \toprule
            \textbf{Step} & \textbf{Acc. (\%)} & \textbf{Help} & \textbf{Rel.} & \textbf{Depth} \\
            \midrule
            0  & 76.97 & 8.81 & 9.55 & 8.32 \\
            2  & 67.88 & 7.08 & 7.48 & 6.76 \\
            4  & 91.52 & 8.08 & 8.47 & 7.15 \\
            6  & 93.94 & 9.58 & 9.83 & 8.81 \\
            8  & 96.36 & 9.38 & 9.61 & 8.55 \\
            10 & 97.58 & 9.33 & 9.62 & 8.51 \\
            12 & 96.36 & 9.52 & 9.78 & 8.80 \\
            14 & 96.97 & 9.73 & 9.91 & 9.01 \\
            16 & 97.58 & {\bf 9.78} & 9.95 & {\bf 9.06} \\
            18 & 97.58 & 9.77 & 9.95 & 9.05 \\
            20 & {\bf 98.18} & 9.77 & {\bf 9.95} & 9.02 \\
            \bottomrule
        \end{tabular}
    \end{minipage}%
    \hfill
    \begin{minipage}{0.52\textwidth}
        \centering
        \begin{tikzpicture}
        \begin{axis}[
            width=\textwidth,
            height=6cm,
            xlabel={Step},
            ylabel={Score},
            ymin=4, ymax=10,
            ytick={5,6,7,8,9,10},
            xmin=0, xmax=20,
            xtick={0,2,4,6,8,10,12,14,16,18,20},
            axis y line*=left,
            axis x line*=bottom,
            ymajorgrids=true,
            enlargelimits=false,
            legend to name=sharedlegendbaseline,
            legend style={draw=gray, font=\small, column sep=1em, /tikz/every even column/.append style={column sep=1em}, legend columns=4, anchor=north},
        ]

        \addplot+[mark=*, thick] coordinates {
            (0,8.81) (2,7.08) (4,8.08) (6,9.58) (8,9.38)
            (10,9.33) (12,9.52) (14,9.73) (16,9.78)
            (18,9.77) (20,9.77)
        };
        \addlegendentry{Helpfulness}

        \addplot+[mark=square*, thick] coordinates {
            (0,9.55) (2,7.48) (4,8.47) (6,9.83) (8,9.61)
            (10,9.62) (12,9.78) (14,9.91) (16,9.95)
            (18,9.95) (20,9.95)
        };
        \addlegendentry{Relevance}

        \addplot+[mark=triangle*, thick] coordinates {
            (0,8.32) (2,6.76) (4,7.15) (6,8.81) (8,8.55)
            (10,8.51) (12,8.80) (14,9.01) (16,9.06)
            (18,9.05) (20,9.02)
        };
        \addlegendentry{Depth}

        \addplot+[mark=diamond*, thick] coordinates {
            (0,7.697) (2,6.788) (4,9.152) (6,9.394) (8,9.636)
            (10,9.758) (12,9.636) (14,9.697) (16,9.758)
            (18,9.758) (20,9.818)
        };
        \addlegendentry{Accuracy}

        \end{axis}

        \begin{axis}[
            width=\textwidth,
            height=6cm,
            xmin=0, xmax=20,
            ymin=40, ymax=100,
            axis y line*=right,
            axis x line=none,
            ylabel={Accuracy (\%)},
            ytick={50,60,70,80,90,100},
            yticklabels={50,60,70,80,90,100},
        ]
        \end{axis}
        \end{tikzpicture}
    \end{minipage}

    \vspace{0.5em}
    \pgfplotslegendfromname{sharedlegendbaseline}
\end{figure}

Figures \ref{figure:baseline-results} and \ref{figure:xml-results} summarize performance trends across training steps. We observe consistent improvements in factual accuracy over time, particularly in the Baseline format. In most cases, gains in accuracy are achieved without sacrificing helpfulness or relevance, and in later checkpoints, all four metrics reach strong levels of performance.

Notably, accuracy rises from 76.97\% at step 0 to 98.18\% at step 20 in the Baseline format, demonstrating the model’s increasing ability to ignore fictitious context. Helpfulness and depth also improve steadily, with a dip at the first generated checkpoint.

\begin{figure}[H]
    \caption{Evaluation results across training steps (XML format). Accuracy is plotted on the right y-axis, and other metrics use the left y-axis.}
    \label{figure:xml-results}
    \centering
    \begin{minipage}{0.45\textwidth}
        \centering
        \begin{tabular}{ccccc}
            \toprule
            \textbf{Step} & \textbf{Acc.} & \textbf{Help} & \textbf{Rel} & \textbf{Depth} \\
            \midrule
            0   & 78.79 & 8.81 & 9.56 & 8.19 \\
            2   & 52.73 & 5.79 & 6.16 & 5.24 \\
            4   & 87.88 & 6.56 & 7.09 & 5.47 \\
            6   & 95.76 & 9.46 & 9.73 & 8.75 \\
            8   & 94.55 & 9.09 & 9.35 & 8.21 \\
            10  & 94.55 & 8.93 & 9.32 & 8.01 \\
            12  & 95.76 & 8.95 & 9.33 & 8.05 \\
            14  & 95.76 & 9.28 & 9.59 & 8.52 \\
            16  & 97.58 & 9.35 & 9.61 & 8.61 \\
            18  & {\bf 97.58} & 9.28 & 9.50 & 8.50 \\
            20  & 96.97 & {\bf 9.40} & {\bf 9.64} & {\bf 8.64} \\
            \bottomrule
        \end{tabular}
    \end{minipage}%
    \hfill
    \begin{minipage}{0.52\textwidth}
        \centering
        \begin{tikzpicture}
        \begin{axis}[
            width=\textwidth,
            height=6cm,
            xlabel={Step},
            ylabel={Score},
            ymin=4, ymax=10,
            ytick={5,6,7,8,9,10},
            xmin=0, xmax=20,
            xtick={0,2,4,6,8,10,12,14,16,18,20},
            axis y line*=left,
            axis x line*=bottom,
            ymajorgrids=true,
            enlargelimits=false,
            legend to name=sharedlegendxml,
            legend style={
                draw=gray,
                font=\small,
                column sep=1em,
                /tikz/every even column/.append style={column sep=1em},
                legend columns=4,
                anchor=north
            },
        ]

        \addplot+[mark=*, thick] coordinates {
            (0,8.81) (2,5.79) (4,6.56) (6,9.46) (8,9.09)
            (10,8.93) (12,8.95) (14,9.28) (16,9.35)
            (18,9.28) (20,9.40)
        };
        \addlegendentry{Helpfulness}

        \addplot+[mark=square*, thick] coordinates {
            (0,9.56) (2,6.16) (4,7.09) (6,9.73) (8,9.35)
            (10,9.32) (12,9.33) (14,9.59) (16,9.61)
            (18,9.50) (20,9.64)
        };
        \addlegendentry{Relevance}

        \addplot+[mark=triangle*, thick] coordinates {
            (0,8.19) (2,5.24) (4,5.47) (6,8.75) (8,8.21)
            (10,8.01) (12,8.05) (14,8.52) (16,8.61)
            (18,8.50) (20,8.64)
        };
        \addlegendentry{Depth}

        \addplot+[mark=diamond*, thick] coordinates {
            (0,7.879) (2,5.273) (4,8.788) (6,9.576) (8,9.455)
            (10,9.455) (12,9.576) (14,9.576) (16,9.758)
            (18,9.758) (20,9.697)
        };
        \addlegendentry{Accuracy}

        \end{axis}

        \begin{axis}[
            width=\textwidth,
            height=6cm,
            xmin=0, xmax=20,
            ymin=40, ymax=100,
            axis y line*=right,
            axis x line=none,
            ylabel={Accuracy (\%)},
            ytick={50,60,70,80,90,100},
            yticklabels={50,60,70,80,90,100},
        ]
        \end{axis}
        \end{tikzpicture}
    \end{minipage}

    \vspace{0.5em}
    \pgfplotslegendfromname{sharedlegendxml}
\end{figure}

\newpage
\subsection{Ablation: Effect of Prompt Structure}
\label{prompt-structure-ablation}

To assess the impact of prompt formatting on hallucination resistance, we perform an ablation study comparing two versions of Finetune-RAG: one trained using the \textbf{Baseline format} and another using a more structured \textbf{XML format}. Both models were fine-tuned on the same dataset with identical hyperparameters and evaluated using the same GPT-4o-based benchmarking pipeline.

\paragraph{Prompt Format Differences} The Baseline format presents context in a flat, unstructured layout, while the XML format uses nested tags to explicitly delineate retrieved content blocks (see Section \ref{user-message}). We hypothesized that structured formatting might help the model better separate and reason about distinct chunks.

\paragraph{Baseline Format}
This format presents the retrieved content in a plain and direct layout:
\begin{verbatim}
Filename: {filename1}
Information:
{content1}

Filename: {filename2}
Information:
{content2}

Question: {question}
\end{verbatim}

\paragraph{XML Format}
This version wraps the content in an XML-like structure for clearer boundaries:
\begin{verbatim}
<Results>
    <Result>
        <Filename>{filename1}</Filename>
        <Information>{content1}</Information>
    </Result>
    <Result>
        <Filename>{filename2}</Filename>
        <Information>{content2}</Information>
    </Result>
</Results>

Question: {question}
\end{verbatim}

\paragraph{Results} As shown in Figures \ref{figure:baseline-results} and \ref{figure:xml-results}, both models demonstrate strong improvements over time. However, the Baseline model consistently achieves higher accuracy and better overall scores in the later checkpoints:

\begin{itemize}
    \item At step 20, the Baseline-tuned model achieves an accuracy of 98.18\%, compared to 96.97\% for the XML-tuned model.
    \item The Baseline-tuned model also maintains slightly higher scores for helpfulness (9.77 vs 9.40) and depth (9.02 vs 8.64).
\end{itemize}

\paragraph{Interpretation} These results suggest that while XML-style formatting introduces clear structural boundaries that aid human readers, it did not consistently outperform the simpler Baseline prompt. We offer two possible explanations: (1) the model may have developed inductive biases from pretraining that favor interpreting flat, plain-text layouts, such as those seen in summaries or abstracts, and (2) fine-tuning datasets used in LLaMA or similar models may have predominantly featured unstructured prompts, making the model more adept at handling them.

This suggests that while prompt formatting is an important factor, training data design and supervision signal play a larger role in hallucination resistance.

\section{Discussion}

Our results show that Finetune-RAG significantly improves a model's ability to resist hallucinations in a RAG setting, even when the prompt includes both correct and misleading context. Fine-tuning with dual-context examples leads to consistent improvements in factual accuracy, while preserving helpfulness, relevance, and depth.

\subsection{Inductive Bias Emergence in Structure-Agnostic Learning}

A significant and perhaps unexpected result in our study is that models trained on unstructured prompts (Baseline format) performed better, especially in factual accuracy, compared to those trained with structured XML prompts. This challenges the common belief that clear structure always aids reasoning. Instead, it suggests a deeper learning process, which involves the development of stronger built-in tendencies for selecting content when structure is absent. This raises a potential area that can be further researched upon.

\subsection{Limitations}

Despite promising results, several limitations remain:

\begin{itemize}
    \item \textbf{Synthetic dataset generation}: The fictitious content is generated using GPT-4o \citep{openai2024gpt4o}, which may introduce distributional artifacts that differ from real-world retrieval errors. Additionally, the size of the dataset can be further increased for effective fine-tuning in larger models.
    
    \item \textbf{Binary supervision}: We treat hallucination as a binary decision at the generation level. However, hallucination is often more nuanced, involving partial truths, omissions, or subtle phrasing, which our current framework may not sufficiently address.

    \item \textbf{Controlled context pairing}: During training, each example includes exactly one correct and one incorrect document chunk. This creates a simplified binary contrast that may not generalize to real-world scenarios where multiple retrieved documents vary in quality. A stronger training approach can be constructed using our existing dataset to create  more varied and robust scenarios that the model can train on.

    \item \textbf{Compute requirements}: While our method is simpler and less resource-intensive than alternatives such as full retraining or reinforcement learning, it still requires access to a high-memory GPU (e.g., H100) to fine-tune long-context models with large batch sizes. This may limit accessibility for some users or institutions.

\end{itemize}

\subsection{Future Work}

There are several promising extensions to Finetune-RAG that could further improve its robustness and applicability:

\begin{itemize}
    \item \textbf{Training with more in-context RAG}: Real-world retrieval often returns more than two documents, and the context window of LLMs are increasing rapidly. At the time of our work, we focused on relatively low context window of 8k, which would realistically be used for two to three RAG documents using up to 3k context window. With increasing context window, future work can explore training with more RAG chunks to optimize LLMs RAG performance even at high level of stresses caused by more retrieved chunks. To support this, we future-proofed our dataset by including two additional relevant chunks per example to support generating more complex multi-document training scenarios.

    \item \textbf{Joint retrieval-generation optimization}: While Finetune-RAG focuses on improving the generation component, combining it with learned retrieval mechanisms such as reranker-aware retrievers or contrastively trained retrievers could lead to further improvements in factual accuracy and context filtering.

    \item \textbf{Multimodal extensions}: Hallucination is not limited to text-based models. Extending Finetune-RAG to multimodal settings, such as image-caption retrieval or code+documentation generation, may help build more robust grounded systems in other domains.

    \item \textbf{Evaluation on downstream tasks}: While our benchmarking focuses on controlled hallucination settings, future work should assess Finetune-RAG's impact on end-to-end performance in downstream RAG applications such as open-domain question answering, legal document summarization, and domain-specific information retrieval.

\end{itemize}

\section{Conclusion}

In this work, we present \textbf{Finetune-RAG}, a simple yet effective method for reducing hallucination in Retrieval-Augmented Generation (RAG) through supervised fine-tuning. Rather than focusing on retrieval quality, Finetune-RAG trains the generation model to rely solely on factual context while ignoring misleading information, with no architectural changes required.

We constructed a diverse training set and evaluate using \textbf{Bench-RAG}, a technique that leverages GPT-4o as an automatic judge. Results show substantial gains in factual accuracy while preserving helpfulness, relevance, and depth. Ablation studies further reveal that prompt structure subtly impacts robustness, with less structured formats sometimes aiding discrimination.

Despite its simplicity, Finetune-RAG demonstrates that generation-stage fine-tuning can meaningfully improve hallucination resistance in noisy retrieval environments. We release our code, dataset, and checkpoints to support further research in this direction, and highlight future extensions including multi-document training, joint retrieval-generation optimization, and adaptation to multimodal tasks.

\newpage
\bibliographystyle{apalike}
\bibliography{main}

\newpage
\appendix

\section{Dataset Example Format}
\label{appendix:dataset-example-format}

Aligning with Section \ref{problem-setup}, each example in the dataset is structured as follows:

\begin{verbatim}
{
  "content": <factual chunk>,
  "filename": <original document filename>,
  "fictitious_content": <misleading chunk>,
  "fictitious_filename": <filename of misleading chunk>,
  "question": <user query>,
  "answer": <GPT-4o generated answer based only on correct content>,
}
\end{verbatim}

\section{Bench-RAG Prompt Structure}
\label{appendix:bench-rag-prompt-structure}

Given both the correct and fictitious document chunks, the fine-tuned model checkpoints are used to generate answers for questions on the test dataset. The outputs are stored in a structured \texttt{jsonl} format, with each entry containing the following fields:

\begin{verbatim}
{
  "filename": <original document filename>,
  "content": <factual chunk>,
  "question": <user query>,
  "response": <model's generated answer>
}
\end{verbatim}

With these output, we curate a prompt for the four measurements derived from our evaluation.

\subsection{System Message for Evaluation}
\label{appendix:evaluation-system-message}

\paragraph{Accuracy}
\begin{verbatim}
"Please act as an impartial judge and evaluate the quality of the
response provided by an AI assistant to the user question displayed
below, based solely on a piece of information extracted from a file
provided below. Your evaluation should consider the accuracy of the
response. 

You will check whether the response contains extra details not found
in the piece of information provided. If extra details are found,
accuracy is false. Otherwise, accuracy is true. Take note that if the
response partially addresses the question, but did not provide extra
details not found in the piece of information provided, the response
will still be considered accurate (hence accuracy = true).

Begin your evaluation by providing a short explanation. Be as objective
as possible. After providing your explanation, you must rate the accuracy
with true or false by strictly following this JSON format:
{
    "accuracy_explanation":
        <provide an explanation on accuracy, whether extra details
        outside the content were found.>,
    "accuracy": <true/false>
}"
\end{verbatim}

\paragraph{Helpfulness}

\begin{verbatim}
"Please act as an impartial judge and evaluate the quality of the
response provided by an AI assistant to the user question displayed
below, based solely on a piece of information extracted from a file
provided below. Your evaluation should consider the helpfulness of the
response. 

You will check whether the AI assistant is helpful in answering the
question based on the response.

Begin your evaluation by providing a short explanation. Be as objective
as possible. After providing your explanation, you must rate the
helpfulness on a scale of 1 to 10 by strictly following this JSON format:
{
    "helpfulness_explanation": <provide an explanation on helpfulness>,
    "helpfulness": <score>
}"
\end{verbatim}

\paragraph{Relevance}

\begin{verbatim}
"Please act as an impartial judge and evaluate the quality of the
response provided by an AI assistant to the user question displayed
below, based solely on a piece of information extracted from a file
provided below. Your evaluation should consider the relevance of the
response. 

You will check the relevance of the response by evaluating whether the
response fully addresses the question.

Begin your evaluation by providing a short explanation. Be as objective
as possible. After providing your explanation, you must rate the
relevance on a scale of 1 to 10 by strictly following this JSON format:
{
    "relevance_explanation": <provide an explanation on relevance>,
    "relevance": <score>
}"
\end{verbatim}

\paragraph{Depth}

\begin{verbatim}
"Please act as an impartial judge and evaluate the quality of the
response provided by an AI assistant to the user question displayed
below, based solely on a piece of information extracted from a file
provided below. Your evaluation should consider the depth of the
response. 

You will check the depth of the response by evaluating the level of
detail of the response in answering the question.


Begin your evaluation by providing a short explanation. Be as objective
as possible. After providing your explanation, you must rate the
depth on a scale of 1 to 10 by strictly following this JSON format:
{
    "depth_explanation": <provide an explanation on depth>,
    "depth": <score>
}"
\end{verbatim}

\subsection{User Message for Evaluation}
\label{appendix:evaluation-user-message}

All measurements utilizes the same user message structure for evaluation. Note that the content used is the correct content, rather than the fictitious one:

\begin{verbatim}
[The Start of Provided Information Extracted from a File]
Filename: {filename}

Information: {content}
[The End of Provided Information]
            
[Question]
{question}

[The Start of Assistant's Response]
{response}
[The End of Assistant's Response]
\end{verbatim}

\end{document}